\documentclass[12pt]{article}
\usepackage{authblk}
\usepackage{setspace}  
\usepackage{graphicx} 
\usepackage{geometry}
\usepackage{xcolor}
\usepackage{comment}
\usepackage{fixltx2e}

\usepackage{multirow}
\usepackage{graphicx} 
\usepackage{soul}

\title{Unlocking the Potential of Generative AI through Neuro-Symbolic Architectures – Benefits and Limitations}
\author[1]{Oualid Bougzime}
\author[2]{Samir Jabbar}
\author[2]{Christophe Cruz}
\author[1,3]{Frédéric Demoly}
\affil[1]{ICB UMR 6303 CNRS, Université Marie et Louis Pasteur, UTBM, 90010 Belfort Cedex, France}
\affil[2]{ICB UMR 6303 CNRS, Université Bourgogne Europe, 21078 Dijon, France}
\affil[3]{Institut universitaire de France (IUF), Paris, France}
\date{}

\usepackage{amsmath}
\usepackage{tikz}
\usepackage{adjustbox}
\usepackage{caption} 
\usetikzlibrary{shapes.geometric, arrows, positioning}

\definecolor{neuralColor}{HTML}{1f77b4}
\definecolor{symbolicColor}{HTML}{ff7f0e}
\definecolor{rootColor}{HTML}{2ca02c}

\tikzset{
  block/.style={rectangle, draw, thick, rounded corners, minimum height=0.5cm, minimum width=2cm, text centered, font=\footnotesize},
  neural/.style={block, fill=neuralColor!30},
  symbolic/.style={block, fill=symbolicColor!30},
  root/.style={block, fill=rootColor!30},
  arrow/.style={thick,->,>=stealth}
}

\usepackage{url} 
\usepackage{breakurl}
\usepackage{hyperref} 
\begin{document}
\maketitle
\doublespacing
\section*{Abstract}
Neuro-symbolic artificial intelligence (NSAI) represents a transformative approach in artificial intelligence (AI) by combining deep learning's ability to handle large-scale and unstructured data with the structured reasoning of symbolic methods. By leveraging their complementary strengths, NSAI enhances generalization, reasoning, and scalability while addressing key challenges such as transparency and data efficiency. This paper systematically studies diverse NSAI architectures, highlighting their unique approaches to integrating neural and symbolic components. It examines the alignment of  contemporary AI techniques such as retrieval-augmented generation, graph neural networks, reinforcement learning, and multi-agent systems with NSAI paradigms. This study then evaluates these architectures against comprehensive set of criteria, including generalization, reasoning capabilities, transferability, and interpretability, therefore providing a comparative analysis of their respective  strengths and limitations. Notably, the Neuro → Symbolic ← Neuro model consistently outperforms its counterparts across all evaluation metrics. This result aligns with state-of-the-art research that highlight the efficacy of such architectures in harnessing advanced technologies like multi-agent systems.

\vspace*{0.5cm}

\noindent \underline{Keywords}: Neuro-symbolic Artificial Intelligence, Neural Network, Symbolic AI, Generative AI, Retrieval-Augmented Generation (RAG), Reinforcement Learning (RL), Natural Language Processing (NLP), Explainable AI (XAI), Benchmark

\section{Introduction}
Neuro-symbolic artificial intelligence (NSAI) is fundamentally defined as the combination of deep learning and symbolic reasoning \cite{garcez2023neurosymbolic}. This hybrid approach aims to overcome the limitations of both symbolic and neural artificial intelligence (AI) systems while harnessing their respective strengths. Symbolic AI excels in reasoning and interpretability, whereas neural AI thrives in learning from vast amounts of data. By merging these paradigms, NSAI aspires to embody two  fundamental aspects of intelligent cognitive behavior: the ability to learn from experience and the capacity to reason based on acquired knowledge \cite{garcez2023neurosymbolic, valiant2003three}.

\vspace*{0.5cm}

The importance of NSAI has been increasingly recognized in recent years, especially after the 2019 Montreal AI Debate between Gary Marcus and Yoshua Bengio. This debate highlighted two contrasting perspectives on the future of AI: Marcus argued that “expecting a monolithic architecture to handle abstraction and reasoning is unrealistic,” emphasizing the limitations of current AI systems, while Bengio maintained that “sequential reasoning can be performed while staying in a deep learning framework” \cite{bengio2019ai}. This discussion brought attention to the strengths and weaknesses of neural and symbolic approaches, catalyzing a surge of interest in hybrid solutions. Bengio’s subsequent remarks at IJCAI 2021 underscored the importance of addressing out-of-distribution (OOD) generalization, stating that “we need a new learning theory” to tackle this critical challenge \cite{bengio2022system}. This aligns with the broader consensus within the AI community that combining neural and symbolic paradigms is essential to developing more robust and adaptable systems. Drawing on concepts like Daniel Kahneman’s dual-process theory of reasoning, which compares fast, intuitive thinking (System 1) to deliberate, logical thought (System 2), NSAI seeks to bridge the gap between learning from data and reasoning with structured knowledge \cite{marcus2019rebooting}. Despite ongoing debates about the optimal architecture for integrating these two paradigms, the 2019 Montreal AI Debate has played a pivotal role in shaping the trajectory of research in this promising field \cite{marcus2018deep, liu2022neural, zhang2021neural, lamb2020graph, von2021informed, belle2020symbolic}.

\vspace*{0.5cm}

NSAI offers a promising avenue for addressing limitations of purely symbolic or neural systems. For instance, while neural networks (NNs) often struggle with interpretability, symbolic AI systems are rigid and require extensive domain knowledge. By combining the adaptability of neural models with the explicit reasoning capabilities of symbolic methods, NSAI systems aim to provide enhanced generalization, interpretability, and robustness. These characteristics make NSAI particularly well-suited for solving complex, real-world problems where adaptability and transparency are critical \cite{hamilton2024neuro}. Several NSAI architectures have been proposed to integrate these paradigms effectively. Examples include Symbolic Neuro Symbolic systems, Symbolic[Neuro], Neuro[Symbolic], Neuro | Symbolic coroutines, Neuro\textsubscript{Symbolic}, and others \cite{kautz2022third}. Each architecture offers unique advantages but also poses specific challenges in terms of scalability, interpretability, and adaptability. A systematic evaluation of these architectures is imperative to understand their potential and limitations, guiding future research in this rapidly evolving field.

\vspace*{0.5cm}

Generative AI has witnessed remarkable advancements, encompassing a diverse range of technologies that address various challenges in data processing, reasoning, and decision-making. These advancements can be categorized into several major branches of AI. Natural language processing (NLP) \cite{vaswani2017attention} includes technologies such as retrieval-augmented generation (RAG) \cite{lewis2020retrieval}, sequence-to-sequence models \cite{sutskever2014sequence}, semantic parsing \cite{jiang2024survey}, named entity recognition (NER) \cite{marrero2013named}, and relation extraction \cite{zhao2024comprehensive}, which focus on understanding and generating human language. Reinforcement learning techniques, like reinforcement learning with human feedback (RLHF) \cite{christiano2017deep}, enable systems to learn optimal actions through interaction with their environment. Advanced NNs include innovations such as graph neural networks (GNNs) \cite{zhou2020graph} and generative adversarial networks (GANs) \cite{goodfellow2014generative}, which excel in handling structured data and generating realistic data samples, respectively. Multi-agent systems \cite{guo2024large, maldonado2024multi} explore the coordination and decision-making among multiple intelligent agents. Recent advances leverage mixture of experts (MoE) architectures to enhance scalability and specialization in collaborative frameworks. In MoE-based multi-agent systems, each expert operates as an autonomous agent, specializing in distinct sub-tasks or data domains, while a dynamic gating mechanism orchestrates their contributions \cite{he2024mixturemillionexperts, lo2024closerlookmixtureofexpertslarge}. Transfer Learning \cite{alyafeai2020survey}, including pre-training \cite{devlin2018bert}, fine-tuning \cite{howard2018universal}, and few-shot learning \cite{parnami2022learning}, allows AI models to adapt knowledge from one task to another efficiently. Explainable AI (XAI) \cite{arrieta2020explainable} focuses on making AI systems transparent and interpretable, while efficient learning techniques, such as model distillation \cite{hinton2015distilling}, aim to optimize resource usage. Reasoning and inference methods like chain-of-thought (CoT) \cite{wei2022chain} reasoning and link prediction enhance logical decision-making capabilities. Lastly, continuous learning \cite{chen2018lifelong} paradigms ensure adaptability over time. Together, these technologies form a comprehensive toolkit for tackling the increasingly complex demands of generative AI applications.

\vspace*{0.5cm}

The classification of generative AI technologies within the NSAI framework is crucial for several reasons. Firstly, it provides a structured approach to understanding how these diverse technologies relate to and enhance NSAI capabilities. By mapping these techniques to specific NSAI architectures, researchers and practitioners can better grasp their potential applications and limitations. This classification also facilitates the identification of synergies between different AI approaches, potentially leading to more robust and versatile hybrid systems. Furthermore, it aids in decision-making processes when selecting appropriate technologies for specific tasks, considering factors like interpretability, reasoning capabilities, and generalization. As AI continues to evolve, this systematic categorization becomes increasingly valuable for bridging the gap between cutting-edge research and practical implementation, ultimately driving the field towards more integrated and powerful AI solutions.

\vspace*{0.5cm}

Therefore, this research aims to explores the alignment of generative AI technologies with the core catergories of NASAI and examines the insights this classification provides regarding their strenghts and limitations. The proposed methodology is threefold: (i) to define and analyze existing NSAI architectures, (ii) to classify generative AI technologies within the NSAI framework to provide a unified perspective on their integration, and (iii) to develop a systematic framework for assessing NSAI architectures across various criteria.

\section{Neuro-Symbolic AI: Combining Learning and Reasoning to Overcome AI's Limitations}

\noindent NNs have been exemplary in handling unstructured forms of data, e.g., images, sounds, and textual data. The capacity of these networks to acquire sophisticated patterns and representations from voluminous datasets has provided major breakthroughs in a series of disciplines, from computer vision, speech recognition, to NLP \cite{kenton2019bert,vaswani2017attention}. One of the major benefits of NNs is that they learn and become better from raw data without requiring pre-coded rules or expert knowledge. This makes them highly scalable and efficient to utilize in applications with large raw data. However, despite these benefits, NNs also have some very well-documented disadvantages. One of the major ones of these might be that they are not transparent. Indeed, neural models pose interpretability challenges, making it difficult to understand the process by which they arrive at specific decisions or predictions. Such opacity causes problems for critical applications where explanation is necessary, such as in healthcare, finance, legal frameworks, and engineering. Additionally, NNs have a high requirement for data, requiring substantial amounts of labeled training data in order to operate effectively. This reliance on large data makes them ineffective when applied to data-scarce or data-costly environments. Neural models also struggle with reasoning and generalizing beyond their training data, which makes their performance less impressive when it comes to tasks in logical inference or commonsense reasoning. Specifically, tasks including understanding causality, sequential problem-solving, and decision-making relying on outside world knowledge.

\vspace*{0.5cm}

Symbolic AI is better at handling areas that are weaker for NNs. Symbolic systems function on explicit rules and structured representations, which enables them to achieve reasoning tasks related to complicated issues, such as mathematical proofs, planning, and expert systems. Symbolic AI is most important because it is transparent. Since symbolic methods are grounded in known rules and logical formalisms, decision-making processes are easy to interpret and explain. However, symbolic AI systems have some drawbacks. One of the biggest ones is that they are rigid and difficult to respond to new circumstances. They require rules to be manually defined and require structured input data, leading them difficult to apply to real-world situations where data might contain noise, incompleteness, or unstructured form. They are also susceptible to combinatorial explosions in handling big data or hard reasoning problems, which significantly slows down their performance at scale. Symbolic systems are also not well suited for perception tasks like image or speech recognition since they are unable to draw knowledge from raw data alone.

\vspace*{0.5cm}

While traditional NNs are strong at recognizing patterns in collections of data but falter when presented with new situations, symbolic reasoning provides a rational foundation for decision-making but is limited in the manner in which it can learn knowledge from new information and adapt in a dynamic process. The combination of these two approaches in NSAI effectively minimizes these limitations, producing a more flexible, explainable, and effective AI system.
Another distinguishing feature of NSAI is that it is able to generalize outside its training set. Traditional AI systems are prone to fail in novel situations; however, NSAI, because of its combination of learning and logical reasoning, works better in such cases. Such a feature is critical for real-world applications such as autonomous transport and medicine, where systems need to perform well in uncontrolled environments. Apart from that, in an interdisciplinary engineering context such as 4D printing, which brings together materials science, additive manufacturing, and engineering, NSAI holds significant promise for improving both the interpretability and reliability of design decisions on the actuation and mechanical performance, and printability. Although these advantages seem promising, they remain hypotheses requiring more extensive validation and industrial-scale testing. Ongoing research must demonstrate, through empirical studies and real-world implementations, how NSAI can reliably accelerate the discovery of smart materials and structures \cite{bougzime2025nsai4d}. The second key benefit point of NSAI is that it has a reduced need for big data sets. Traditional AI systems usually require a tremendous amount of data to operate, which might be very time- and resource-consuming. NSAI, however, is able to do better with a much smaller set of data required, due to its symbolic reasoning ability. This makes it a more sustainable and viable option, especially for small organizations or new research areas with limited resources. Along with the aforementioned data efficiency, NSAI models also have the exceptional transferability, i.e., their capacity for using knowledge learned from one task and applying it in another with less need for retraining. Such a property is highly desirable in situations where there is little data related to a new task.

\vspace*{0.5cm}


\section{Neuro-Symbolic AI Architectures}
This section provides an overview of various NSAI architectures, offering insights into their design principles, integration strategies, and unique capabilities. While Kautz’s classification \cite{kautz2022third} serves as a foundational framework, we extend it by incorporating additional architectural perspectives to capture the evolving landscape of NSAI systems. These approaches range from symbolic systems augmented by neural modules for specialized tasks to deeply integrated models where explicit reasoning engines operate within neural frameworks. This expanded categorization highlights the diversity of design strategies and the broad applicability of NSAI techniques, emphasizing their potential for more interpretable, robust, and data-efficient AI solutions.

\subsection{Sequential}
\begin{figure}[h!]
    \centering
    \includegraphics[width=0.6\linewidth]{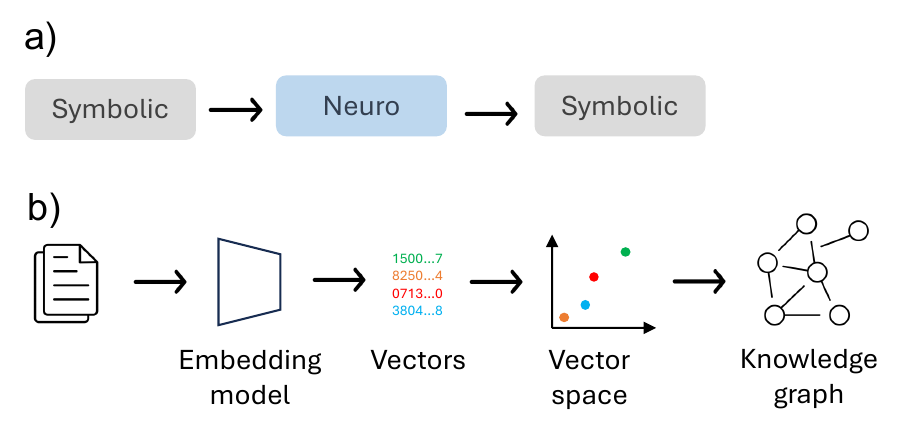}
    \caption{Sequential architecture: (a) Principle and (b) application to knowledge graph construction.}
    \label{fig:sequential}
\end{figure}

As part of the sequential NSAI, the \textit{Symbolic} $\to$ \textit{Neuro} $\to$ \textit{Symbolic} architecture involves systems where both the input and output are symbolic, with a NN acting as a mediator for processing (Figure~\ref{fig:sequential}a). Symbolic input, such as logical expressions or structured data, is first mapped into a continuous vector space through an encoding process. The NN operates on this encoded representation, enabling it to learn complex transformations or patterns that are difficult to model symbolically. Once the processing is complete, the resulting vector is decoded back into symbolic form, ensuring that the final output aligns with the structure and semantics of the input domain. This framework is especially useful for tasks that require leveraging the generalization capabilities of NNs while preserving symbolic interpretability \cite{lample2019deep, dimassi2021ontology}. A formulation of this architecture is presented below:

\begin{equation}
y = f_\text{neural}(x)   
\end{equation}

\noindent where $x$ is the symbolic input, $f_\text{neural}(x)$ represents the NN that processes the input, and $y$ is the symbolic output. 

This architecture can be used in a semantic parsing task, where the input is a sequence of symbolic tokens (e.g., words). Here, each token is mapped to a continuous embedding via word2vec, GloVe, or a similar method \cite{mikolov2013efficient, pennington2014glove}. The NN then processes these embeddings to learn compositional patterns or transformations. From this, the network’s output layer decodes the processed information back into a structured logical form (such as knowledge-graph triples), as illustrated in Figure~\ref{fig:sequential}b.


\subsection{Nested}
\begin{figure}[!h]
    \centering
    \includegraphics[width=0.75\linewidth]{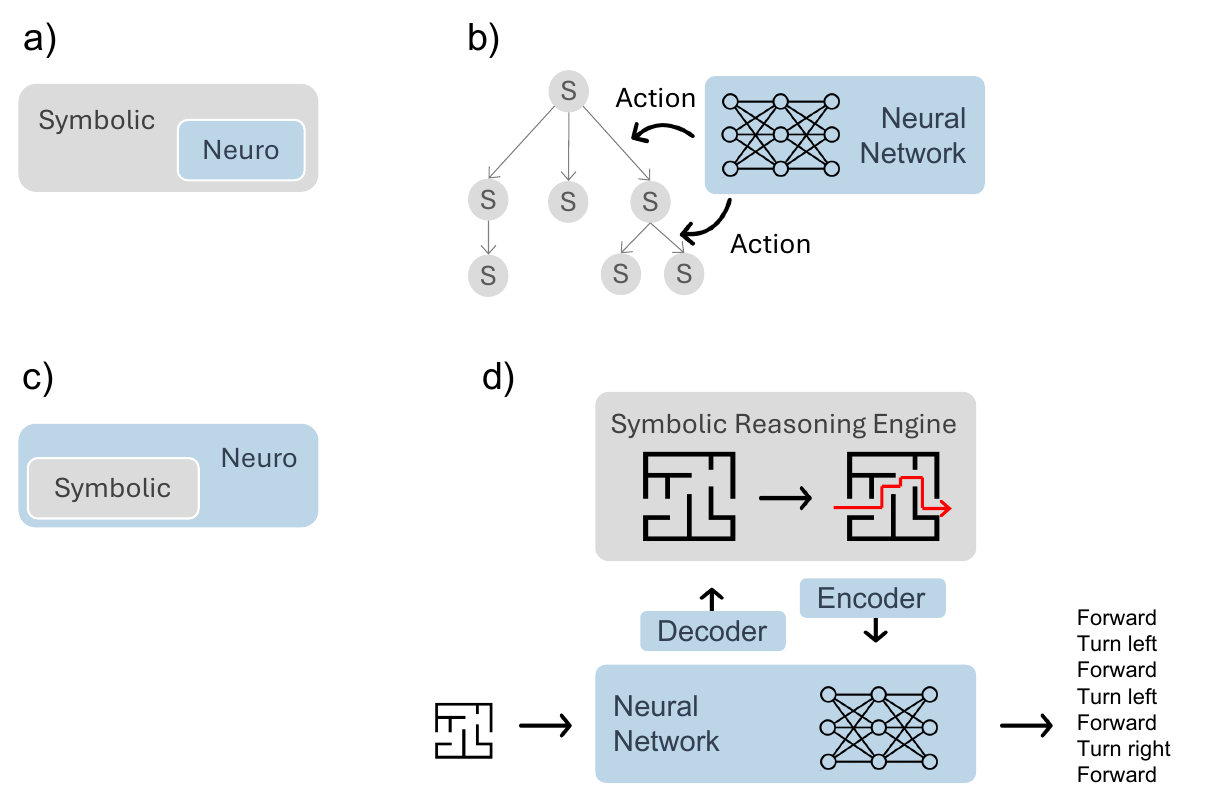}
    \caption{Nested architectures: (a) \textit{Symbolic[Neuro]} principle and (b) its application to tree Search, (c) \textit{Neuro[Symbolic]} principle and (d) its application to maze-solving.}
    \label{fig:nested}
\end{figure}

The nested NSAI category is composed of two different architectures. The first -- \textit{Symbolic[Neuro]} -- places a NN as a subcomponent within a predominantly symbolic system (Figure~\ref{fig:nested}a). Here, the NN is used to perform tasks that require statistical pattern recognition, such as extracting features from raw data or making probabilistic inferences, which are then utilized by the symbolic system. The symbolic framework orchestrates the overall reasoning process, incorporating the neural outputs as intermediate results \cite{silver2016mastering}. This architecture can formally defined as follows:

\begin{equation}
y = g_\text{symbolic}(x, f_\text{neural}(z))  
\end{equation}

\noindent where $x$ represents the symbolic context, $z$ is the input passed from the symbolic reasoner to the NN, $f_\text{neural}(z)$ expresses the neural model processing the input, and $g_\text{symbolic}$  the symbolic reasoning engine that integrates neural outputs. A well-known instance of this architecture is AlphaGo \cite{silver2016mastering}, where a symbolic Monte-Carlo tree search orchestrates high-level decision-making, while a NN evaluates board states, providing a data-driven heuristic to guide the symbolic search process \cite{coulom2006efficient} (Figure~\ref{fig:nested}b). Similarly, in a medical diagnosis scenario, a rule-based engine oversees the core diagnostic process by applying expert guidelines to patient history, symptoms, and lab results. At the same time, a NN interprets unstructured radiological images, delivering key indicators such as tumor likelihood. The symbolic system then integrates these indicators into its final decision, combining transparent and rule-driven logic with robust pattern recognition.

The second architecture -- \textit{Neuro[Symbolic]} --  integrates a symbolic reasoning engine as a component within a neural system, allowing the network to incorporate explicit symbolic rules or relationships during its operation (Figure~\ref{fig:nested}c). The symbolic engine provides structured reasoning capabilities, such as rule-based inference or logic, which complement the NN’s ability to generalize from data. By embedding symbolic reasoning within the neural framework, the system gains interpretability and structured decision-making while retaining the flexibility and scalability of neural computation. This integration is particularly effective for tasks that require reasoning under constraints or adherence to predefined logical frameworks \cite{heule2016solving, madan2021fast}. This configuration can be described as follows:

\begin{equation}
 y = f_\text{neural}(x, g_\text{symbolic}(z))   
\end{equation}

\noindent where $x$ represents the input data to the neural system, $z$ is the input passed from the NN to the symbolic reasoner, $g_\text{symbolic}$ is the symbolic reasoning function, and $f_\text{neural}$ denotes the NN processing the combined inputs.

This architecture is currently applied in automated warehouse, where a robot navigates dynamically changing aisles. During normal operation, it relies on a neural policy to select routes based on learned patterns. When it encounters an unexpected obstacle, it offloads route computation to a symbolic solver (e.g., a pathfinding or constraint-satisfaction algorithm), which returns an alternative path. The solver’s output is then integrated back into the neural policy, and the robot resumes its usual pattern-based navigation. Over time, the robot also learns to identify which challenges call for the symbolic solver, effectively blending fast pattern recognition with precise combinatorial planning.

\noindent Figure~\ref{fig:nested}d illustrates this framework, a symbolic reasoning engine processes structured data, such as a maze, to generate a solution path. A NN  encodes the problem into a latent representation and decodes it into a symbolic sequence of actions (e.g., forward, turn left, turn right).

\subsection{Cooperative}
\begin{figure}[!h]
    \centering
    \includegraphics[width=0.75\linewidth]{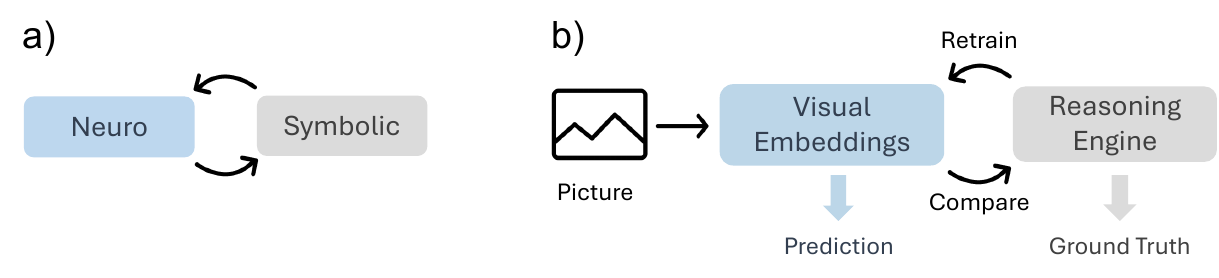}
    \caption{Cooperative architecture: (a) principle and (b) application to visual reasoning.}
    \label{fig:cooperative}
\end{figure}

As a cooperative framework, \textit{Neuro $|$ Symbolic} uses neural and symbolic components as interconnected coroutines, collaborating iteratively to solve a task (Figure~\ref{fig:cooperative}a). NNs process unstructured data, such as images or text, and convert it into symbolic representations that are easier to reason about. The symbolic reasoning component then evaluates and refines these representations, providing structured feedback to guide the NN’s updates. This feedback loop continues over multiple iterations until the system converges on a solution that meets predefined symbolic constraints or criteria. By combining the strengths of NNs for generalization and symbolic reasoning for interpretability, this approach achieves robust and adaptive problem-solving \cite{mao2019neuro}. This architecture can be described as follows:

\begin{equation}
z^{(t+1)} = f_\text{neural}(x, y^{(t)}), \quad y^{(t+1)} = g_\text{symbolic}(z^{(t+1)}), \quad \forall t \in \{0, 1, \dots, n\}    
\end{equation}

\noindent where $x$ represents non-symbolic data input, $z^{(t)}$ is the intermediate symbolic representation at iteration $t$, $y^{(t)}$ is the symbolic reasoning output at iteration $t$, $f_\text{neural}(x, y^{(t)})$ expresses the NN that processes the input $x$ and feedback from the symbolic output $y^{(t)}$, $g_\text{symbolic}(z^{(t+1)})$ is the symbolic reasoning engine that updates $y^{(t+1)}$ based on the neural output $z^{(t+1)}$, and $n$ is the maximum number of iterations or a convergence threshold. The hybrid reasoning halts when the outputs $y^{(t)}$ converge (e.g., $|y^{(t+1)} - y^{(t)}| < \epsilon$)), where $\epsilon$ is a small threshold denoting minimal change between successive outputs, or when the maximum iterations $n$ is reached.

For instance, this architecture can applied in autonomous driving systems, where a NN processes real-time images from vehicle cameras to detect and classify traffic signs. It identifies shapes, colors, and patterns to suggest potential signs, such as speed limits or stop signs. A symbolic reasoning engine then evaluates these detections based on contextual rules—like verifying if a detected speed limit sign matches the road type or if a stop sign appears in a logical position (e.g., near intersections). If inconsistencies are detected, such as a stop sign identified in the middle of a highway, the symbolic system flags the issue and prompts the neural network to re-evaluate the scene. This iterative feedback loop continues until the system reaches consistent, high-confidence decisions, ensuring robust and reliable traffic sign recognition, even in challenging conditions like poor lighting or partial occlusions (Figure~\ref{fig:cooperative}b).

\subsection{Compiled}
\begin{figure}[!h]
    \centering
    \includegraphics[width=1\linewidth]{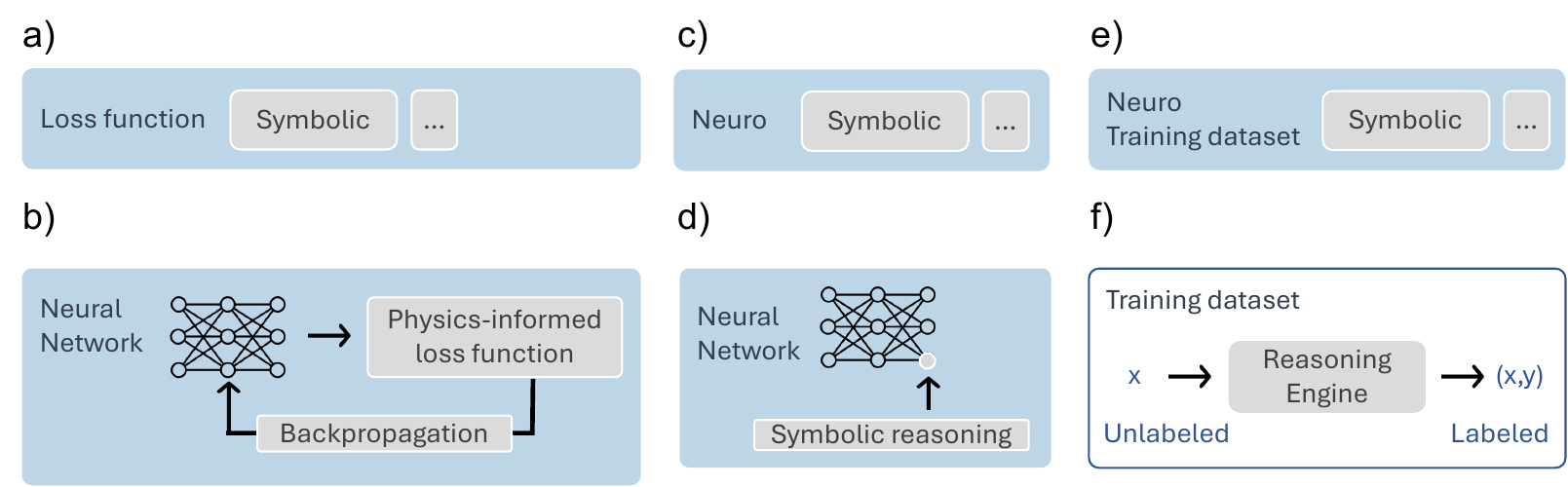}
    \caption{Compiled architectures: (a) \textit{Neuro\textsubscript{Symbolic\textsubscript{Loss}}} principle and (b) application to physics-informed learning; (c) \textit{Neuro\textsubscript{Symbolic\textsubscript{Neuro}}} principle and (d) application of symbolic reasoning in NNs; (e) \textit{Neuro:Symbolic $\rightarrow$ Neuro} principle and (f) application to data Llabeling.}
    \label{fig:compiled}
\end{figure}

As part of the compiled NSAI, \textit{Neuro\textsubscript{Symbolic\textsubscript{Loss}}} uses symbolic reasoning into the loss function of a NN (Figure~\ref{fig:compiled}a). The loss function is typically used to measure the discrepancy between the model's predictions and the true outputs. By incorporating symbolic rules or constraints, the network’s training process not only minimizes prediction error but also ensures that the output aligns with symbolic logic or predefined relational structures. This allows the model to learn not just from data but also from symbolic reasoning, helping to guide its learning process toward solutions that are both accurate and consistent with symbolic principles.

\begin{equation}
 \mathcal{L} = \mathcal{L}_\text{task}(y, y_\text{target}) + \lambda \cdot \mathcal{L}_\text{symbolic}(y)   
\end{equation}

\noindent where $y$ is the model prediction,$y_\text{target}$ represents the ground truth labels, $\mathcal{L}_\text{task}$ is the task-specific loss (e.g., cross-entropy), $\mathcal{L}_\text{symbolic}$ is the penalization for violating symbolic rules, $\lambda$ the Weight balancing the two loss components, and $\mathcal{L}$ the final loss, combining both the task-specific loss and the symbolic constraint penalty to guide model optimization. This architecture is typically useful in the field of 4D printing, where structures need to be optimized at the material level to achieve a target shape. In such a case, a NN predicts the  material distribution and geometric configuration that allows the structure to adapt under external stimuli. The training process incorporates a physics-informed loss function, where, in addition to minimizing the difference between predicted and desired mechanical behavior, the model is penalized whenever the predicted deformation violates symbolic mechanical constraints, such as equilibrium equations or the stress-strain relationship (Figure~\ref{fig:compiled}b). By embedding these symbolic equations directly into the loss function, the NN learns to generate designs that are not only data-driven but also physically consistent, ensuring that the final 4D-printed structure maintains the desired shape across different operational conditions.

A second compiled NSAI architecture, called \textit{Neuro\textsubscript{Symbolic\textsubscript{Neuro}}}, uses symbolic reasoning at the neuron level by replacing traditional activation functions with mechanisms that incorporate symbolic reasoning (Figure~\ref{fig:compiled}c). Rather than using standard mathematical operations like ReLU or sigmoid, the neuron activation is governed by symbolic rules or logic. This allows the NN to reason symbolically at a more granular level, integrating explicit reasoning steps into the learning process. This fusion of symbolic and neural operations enables more interpretable and constrained decision-making within the network, enhancing its ability to reason in a structured and rule-based manner while retaining the flexibility of neural computations. This architecture can be described as follows:

\begin{equation}
 y = g_\text{symbolic}(x)   
\end{equation}

\noindent where: $x$ represents the pre-activation input, $g_\text{symbolic}(x)$ is the symbolic reasoning-based activation function, and $y$ the final neuron. This architecture can find application in lean approval systems, where neural activations are driven by symbolic financial rules rather than traditional functions. One example is the collateral-based constraint neuron, which dynamically adjusts the risk score based on the value of the pledged collateral. When the collateral’s value falls below a predefined threshold relative to the loan amount, the neuron applies a strict penalty that substantially increases the risk score, effectively preventing the system from underestimating the associated financial risk. This symbolic constraint ensures that, regardless of favorable patterns identified in other data, the model consistently accounts for the critical impact of insufficient collateral, leading to more reliable and regulation-compliant credit decisions (Figure~\ref{fig:compiled}d).

Finally, the last compiled architecture, \textit{Neuro:Symbolic $\to$ Neuro}, uses a  symbolic reasoner to generate labeled data pairs \((x, y)\), where \(y\) is produced by applying symbolic rules or reasoning to the input \(x\) (Figure~\ref{fig:compiled}e). These pairs are then used to train a NN, which learns to map from the symbolic input \(x\) to the corresponding output \(y\). The symbolic reasoner acts as a supervisor, providing high-quality, structured labels that guide the NN’s learning process \cite{riegel2020logical}. This architecture can be governed as follows:

\begin{equation}
 \mathcal{D}_\text{train} = \{(x, g_\text{symbolic}(x)) \mid x \in \mathcal{X}\}   
\end{equation}

\noindent where $\mathcal{D}_\text{train}$ is the training dataset, $x$ denotes the unlabeled data, $g_\text{symbolic}(x)$ represents symbolic rules generating labeled data, and $\mathcal{X}$ the set of all input data (Figure~\ref{fig:compiled}b).

Figure~\ref{fig:compiled}f illustrates this architecture, where a reasoning engine is used to label unlabeled training data, transforming raw inputs into structured $(x,y)$ pairs, where symbolic rules enhance the data quality.

\subsection{Ensemble}
\begin{figure}[!h]
    \centering
    \includegraphics[width=0.5\linewidth]{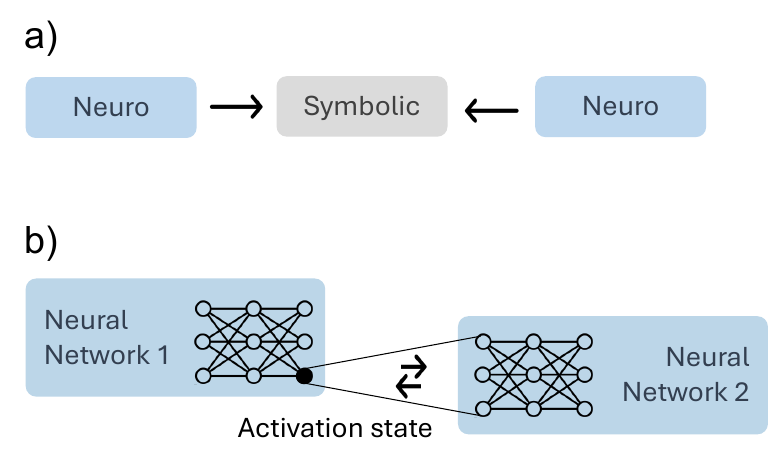}
    \caption{Ensemble architecture: (a) principle and (b) application to NN collaboration.}
    \label{fig:ensemble}
\end{figure}

Another promising architecture, called \textit{Neuro $\to$ Symbolic $\leftarrow$ Neuro} uses multiple interconnected NNs  via a symbolic fibring function, which enables them to collaborate and share information while adhering to symbolic constraints (Figure~\ref{fig:ensemble}a). The symbolic function acts as an intermediary, facilitating communication between the networks by ensuring that their interactions respect predefined symbolic rules or structures. This enables the networks to exchange information in a structured manner, allowing them to jointly solve problems while benefiting from both the statistical learning power of NNs and the logical constraints imposed by the symbolic system \cite{garcez2004fibring}. This architecture can formally defined as follows:

\begin{equation}
 y = g_\text{fibring}(\{f_i\}_{i=1}^n)   
\end{equation}

\noindent where $f_i$ represents the individual NN, $g_\text{fibring}$ is the logic-aware aggregator that enforces symbolic constraints while unifying the outputs of multiple NNs, $n$ the umber of NNs, and $y$ is the combined output of interconnected NNs, produced through the symbolic fibring function $g_\text{fibring}$. For instance in smart cities and urban planning, multiple NNs can be employed, each handle a different urban data stream—such as real-time traffic flow, energy consumption, and air quality measurements. A symbolic fibring function then harmonizes these outputs, enforcing city-level constraints (e.g., ensuring pollution alerts match local environmental regulations, verifying that traffic predictions align with current road network rules). If one network forecasts a surge in vehicle congestion that would push pollution levels beyond acceptable thresholds, the symbolic aggregator identifies the conflict and directs all networks to converge on a coordinated strategy—such as adjusting traffic signals or advising public transport usage. By leveraging each network’s specialized insight within logical urban-planning constraints, the system delivers efficient, consistent decisions across the city’s complex infrastructure.

Figure~\ref{fig:ensemble}b illustrates this architecture, where two NNs (Neural Network 1 and Neural Network 2) communicate through activation states, which enables dynamic exchange of learned representations.

\section{Leveraging NSAI in AI Technologies}

Generative AI is advancing at a remarkable pace, addressing increasingly complex challenges through the integration of diverse methodologies. A key development is the combination of NNs with symbolic reasoning, resulting in hybrid systems that leverage both strengths. Recent studies have demonstrated the effectiveness of this approach in various applications, including design generation and enhancing instructability in generative models \cite{sheth2024neurosymbolic, jacobson2025integrating}.  This section aims to classify  contemporary AI techniques such as RAG, GNNs,  agent-based AI, and transfer learning within the NSAI framework. This classification clarifies how generative AI aligns with neuro-symbolic approaches, bridging cutting-edge research with established paradigms. It also reveals how generative AI increasingly embodies both neural and symbolic characteristics, moving beyond siloed methods.

Additionally, this classification enhances our understanding of these techniques’ roles in AI’s broader landscape, particularly in addressing challenges like interpretability, reasoning, and generalization. It identifies synergies between methods, fostering robust hybrid models that combine neural learning’s adaptability with symbolic reasoning’s precision. Lastly, it supports informed decision-making, guiding researchers and practitioners in selecting the most suitable AI techniques for specific tasks.

\subsection{Overview of Key AI Technologies}

One of the most significant advancements is RAG, which integrates information retrieval with generative models to perform knowledge-intensive tasks. By combining a retrieval mechanism to extract relevant external data with Seq2Seq models for generation \cite{yin2022seq2seq}, RAG excels in applications such as question answering and knowledge-driven conversational AI~\cite{yang2024rag}. Seq2Seq models themselves, built as encoder-decoder architectures, have been pivotal in machine translation, text summarization, and conversational modeling, providing the foundation for many generative AI systems. An extension of RAG is the GraphRAG approach \cite{edge2024local}, which incorporates graph-based reasoning into the retrieval and generation process. By leveraging knowledge graph (KGq) and ontologies structures to represent relationships between information elements, GraphRAG enhances query-focused summarization and reasoning tasks \cite{chen2020review, antoniou2009web}. This method has demonstrated success in producing coherent and contextually rich summaries by integrating local and global reasoning.

GNNs \cite{mavromatis2024gnn} represent a breakthrough in extending neural architectures to graph-structured data, enabling advanced reasoning over interconnected entities. Their ability to model relationships between entities makes them indispensable for a range of tasks, including link prediction, node classification, and recommendation systems, with notable success in KG reasoning. GNNs have also proven highly effective in named entity recognition (NER) \cite{roy2021recent}, where they can leverage graph representations to capture contextual dependencies and relationships between entities in text. This capability extends to relation extraction \cite{wu2024towards}, where GNNs identify and classify semantic relationships between entities, crucial for building and enhancing KG. 

Advances in agentic AI systems, which leverage Large Language Models (LLMs), have shown significant potential in enabling autonomous decision-making and task execution. These systems are designed to function independently, interacting with environments, coordinating with other agents, and adapting to dynamic situations without human intervention. A notable example is AutoGen \cite{wu2023autogen}, a framework that enables the creation of autonomous agents that can interact with each other to solve tasks and improve through continual interactions. Recent work has further enhanced these systems through MoE architectures, which integrate specialized sub-models (``experts") into multi-agent frameworks to optimize task-specific performance and computational efficiency. For instance, MoE-based coordination allows agents to dynamically activate subsets of experts based on context, enabling scalable specialization in complex environments \cite{shazeer2017outrageouslylargeneuralnetworks, lepikhin2020gshard}. Xie et al. \cite{xie2024large} explored the role of LLMs in these agentic systems, discussing their ability to facilitate autonomous cooperation and communication between agents in complex environments, and marking an important step toward scalable and self-sufficient AI. By combining MoE principles with multi-agent collaboration, systems can achieve hierarchical decision-making: LLMs act as meta-controllers, routing tasks to specialized agents (e.g., vision, planning, or language experts) while maintaining global coherence.

However, the growing autonomy of such systems underscores the importance of XAI \cite{ding2022explainability} to ensure transparency and trust. XAI has gained prominence as a means to enhance accountability and support ethical AI adoption. By providing insights into model behavior, XAI ensures that even highly autonomous systems remain interpretable and accountable, addressing concerns about their decisions and actions in sensitive and dynamic environments.

Recent advancements in AI have demonstrated the potential of integrating fine-tuning, distillation, and in-context learning to enhance model performance. Huang et al. \cite{huang2022context} introduced in-context learning distillation, a novel method that transfers few-shot learning capabilities from large pre-trained LLMs to smaller models. By combining in-context learning objectives with traditional language modeling, this approach allows smaller models to perform effectively with limited data while maintaining computational efficiency.

Transfer learning \cite{iman2023review} has similarly emerged as a foundational technique, enabling pre-trained models to adapt their extensive knowledge to new domains using minimal data. This capability is particularly advantageous in resource-constrained scenarios. Techniques such as feature extraction, where pre-trained model layers are repurposed for specific tasks, and fine-tuning, which involves adjusting the weights of the pre-trained model for new tasks, further illustrate its adaptability. 

Complementing these methods, prompt engineering empowers LLMs to perform task-specific functions through carefully designed prompts. Techniques such as CoT prompting \cite{wei2022chain}, zero-shot \cite{pourpanah2022review}, and few-shot prompting  enhance the ability of LLMs to reason and generalize across diverse tasks without extensive retraining \cite{reynolds2021prompt}. Additionally, knowledge distillation  plays a crucial role in optimizing AI models by transferring knowledge from larger, more complex models to smaller, efficient ones \cite{gou2021knowledge}. Variants of distillation, such as task-specific distillation, feature distillation, and response-based distillation, further streamline the process for edge computing and resource-limited environments.

Reinforcement learning and its variant RLHF \cite{dai2023safe}, focus on training agents to make sequential decisions in dynamic environments. RLHF further aligns agent behavior with human preferences, fostering ethical and adaptive AI systems. Finally, continuous learning, or lifelong learning, addresses the challenge of adapting AI systems to new data while retaining previously learned knowledge, ensuring AI remains effective in changing environments \cite{riseAI}.

These techniques represent the cutting edge of generative AI, each contributing to solving complex challenges across diverse applications. The classification of these methods within NSAI paradigm, explored in the following sections, offers a structured perspective on their synergies and practical relevance.

\subsection{Classification of AI Technologies within NSAI Architectures}
\begin{figure}[!h]
    \centering
    \includegraphics[width=0.7\linewidth]{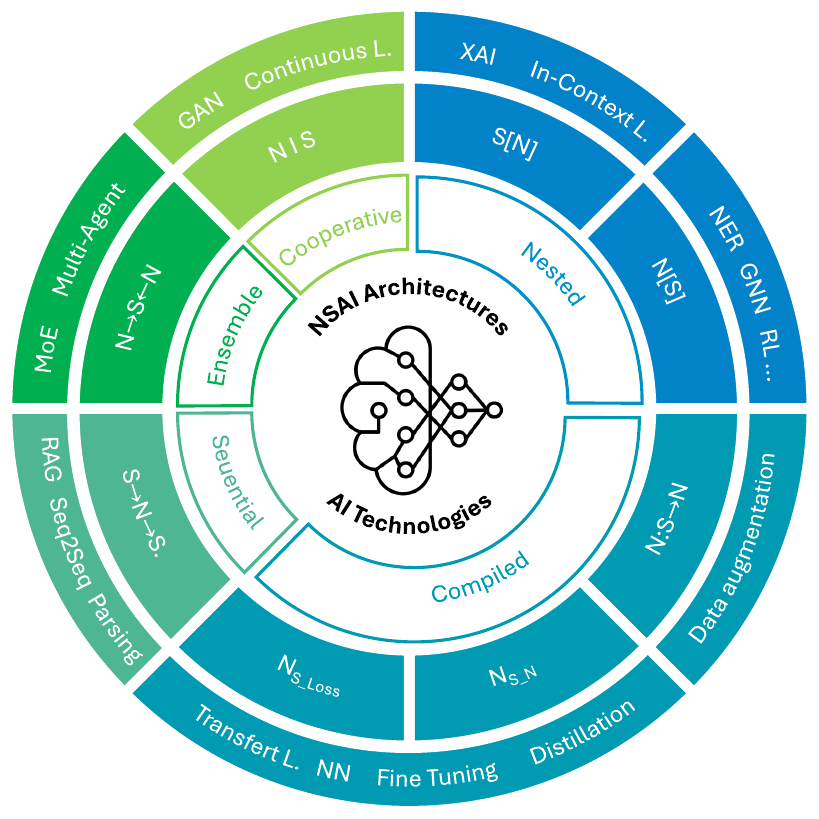}
    \caption{Classification of AI technologies into NSAI architectures.}
    \label{fig:archi}
\end{figure}

This section categorizes generative AI techniques within the eight distinct NSAI architectures, highlighting their underlying principles and practical applications. By classifying these approaches, we gain a clearer understanding of how neural and symbolic methods synergize to address diverse challenges in AI, as summarized in Figure~\ref{fig:archi}.

\subsubsection{The Sequential Paradigm: From Symbolic to Neural Reasoning}
Techniques like RAG, GraphRAG, and Seq2Seq models (including LLMs, e.g., GPT \cite{openai2024gpt4technicalreport}) align with this method due to their reliance on neural encodings of symbolic data (e.g., text or structured information) to perform complex transformations before outputting results in symbolic form. Similarly, semantic parsing benefits from this framework by leveraging NNs to uncover latent patterns in symbolic inputs and generating interpretable symbolic conclusions. For instance, RAG-Logic  proposes a dynamic example-based framework using RAG to enhance logical reasoning capabilities by integrating relevant, contextually appropriate examples \cite{anonymous2024raglogic}. It first encodes symbolic input (e.g., logical premises) into neural representations using the RAG knowledge base search module. Neural processing occurs through the translation module, which transforms the input into formal logical formulas. Finally, the fix module ensures syntactic correctness, and the solver module evaluates the logical consistency of the formulas, decoding the results back into symbolic output. This process maintains the interpretability of symbolic reasoning while leveraging the power of NNs to improve flexibility and performance.

\subsubsection{The Nested Paradigm: Embedding Symbolic Logic in Neural Systems}
In-context learning, such as few-shot learning and CoT reasoning, aligns with the \textit{Symbolic[Neuro]} approach by leveraging NNs for context-aware predictions, while symbolic systems facilitate higher-order reasoning. Similarly, XAI falls into this category, as it often combines neural models for extracting features with symbolic frameworks to produce explanations that are easily understood by humans.

Zhang et al. \cite{zhang2022impact} presented a framework in which symbolic reasoning is enhanced by NNs. CoT is used as a method to generate prompts that combine symbolic rules with neural reasoning. For example, the task of reasoning about relationships between entities, such as “Joseph’s sister is Katherine” is approached by generating a reasoning path through CoT. The reasoning path is structured using symbolic rules, such as $Sister(A,C) \leftarrow Brother(A,B) \land Sister(B,C)$, which define the relationships between entities. These rules are then used to form CoT prompts that guide the model through the reasoning steps. The NN processes these prompts, performing feature extraction and probabilistic inference, while the symbolic system (including the knowledge base and logic rules) orchestrates the overall reasoning process. In this approach, the symbolic framework is the primary system for structuring the reasoning task, and the NN acts as a subcomponent that processes raw data and interprets the symbolic rules in the context of the query.

Methods like GNNs, NER, link prediction, and relation extraction fit into the \textit{Neuro[Symbolic]} category. These methods often leverage symbolic relationships, such as ontologies or graphs, as integral components to enhance neural processing. In addition, they integrate symbolic reasoning subroutines to perform higher-order logical operations, enforce consistency, or derive insights from structured representations. RL and RLHF exemplify this approach, where symbolic reasoning is integrated into the reward shaping and policy optimization stages to enforce logical constraints, ensure decision-making consistency, and align neural outputs with human-like decision-making criteria. For instance, NeuSTIP \cite{singh2023neustip} exemplifies this approach by combining GNN-based neural processing with symbolic reasoning to tackle link prediction and time interval prediction in temporal knowledge graphs (TKGs). NeuSTIP employs temporal logic rules, extracted via ``all-walks" on TKGs, to enforce consistency and strengthen reasoning. By embedding symbolic reasoning subroutines into the neural framework, NeuSTIP demonstrates how such models can effectively derive structured insights and perform reasoning under constraints.

\subsubsection{The Cooperative Paradigm: Iterative Interaction Between Neural and Symbolic Modules}
GANs align with this paradigm as their iterative interplay mirrors a cooperative dynamic between two distinct components: the generator creates outputs, while the discriminator evaluates them against predefined criteria, providing structured feedback to improve the generator's performance. This iterative feedback loop exemplifies the \textit{Neuro $|$ Symbolic} framework, where neural networks and symbolic reasoning components collaborate to achieve robust and adaptive problem-solving while adhering to symbolic constraints or logical consistency. Moreover, this cooperative dynamic inherently facilitates continuous learning, a process in which both neural and symbolic modules undergo iterative refinement to enhance their performance over time. In this paradigm, NN continuously updates its internal representations and model parameters in response to feedback derived from the symbolic module’s logical inferences and constraint evaluations. This adaptive process enables the NN to generalize more effectively across diverse and evolving data distributions. Simultaneously, the symbolic module is not static; it dynamically revises its rule-based reasoning mechanisms and knowledge structures by integrating new information extracted from the NN’s learned representations.
An example of this approach in reinforcement learning is the detect-understand-act (DUA) framework \cite{mitchener2022detect}, where neural and symbolic components collaborate iteratively to solve tasks in a structured manner. In DUA, the detect module uses a traditional computer vision object detector and tracker to process unstructured environmental data into symbolic representations. The understand component, which integrates symbolic reasoning, processes this data using answer set programming (ASP) and inductive logic programming (ILP), ensuring that decisions align with symbolic rules and constraints. The act component, composed of pre-trained reinforcement learning policies, acts as a feedback loop to refine the symbolic representations, allowing the system to converge on solutions that meet predefined criteria.

\subsubsection{The Compiled Paradigm: Embedding Symbolic Reasoning Within Neural Computation}
Approaches such as model distillation, fine-tuning, pre-training, and transfer learning align with the \textit{Neuro\textsubscript{Symbolic}} approach by integrating symbolic constraints or objectives (e.g., logical consistency, relational structures) directly into the learning process of NNs, either through the loss function or at the neuron level via activation functions. This ensures that outputs adhere to predefined symbolic rules, enabling structured reasoning within the network. Consequently, all NN models can be modeled by this paradigm, by embedding symbolic logic into neural architectures, bridging data-driven learning with symbolic reasoning. Mendez-Lucero et al. \cite{mendez2024semantic} complemented this perspective by embedding logical constraints within the loss function. The authors propose a distribution-based method that incorporates symbolic logic, such as propositional formulas and first-order logic, into the learning process. These constraints are encoded as a distribution and incorporated into the optimization procedure using measures like the Fisher-Rao distance or Kullback-Leibler divergence, effectively guiding the NN to adhere to symbolic constraints. This integration of symbolic knowledge into the loss function ensures that the neural model not only learns from data but also incorporates predefined logical rules, reinforcing the connection between neural learning and symbolic reasoning in the context of model distillation, fine-tuning, pre-training, and transfer learning.

Data augmentation leverages the \textit{Neuro:Symbolic $\to$ Neuro} approach, which uses symbolic reasoning to generate synthetic examples, enabling effective data augmentation. By producing high-quality labeled data through logical inference, it enhances the training process of NNs. This method seamlessly integrates the precision and structure of symbolic logic with the scalability and adaptability of NNs, resulting in more robust and efficient learning.
Li et al. \cite{li2024neuro} proposed a methodological framework that exemplifies this approach. Their framework systematically generates labeled data pairs \((x, y)\), where \(y\) is derived from \(x\) through symbolic transformations based on formal logical rules. The process begins with the formalization of mathematical problems in a symbolic space using mathematical solvers, ensuring the logical validity of the generated instances. Subsequently, mutation mechanisms are applied to diversify the examples, including simplification strategies (reducing the complexity of expressions) and complication strategies (adding constraints or variables). Each transformation results in a new problem instance with its corresponding solution, forming labeled pairs \((x', y')\) that enrich the training corpus with controlled complexity levels.

\subsubsection{The Fibring Paradigm: Connecting Neural Models Through Symbolic Constraints}
Techniques such as multi-agent AI and MoE systems align with this paradigm by leveraging symbolic functions to facilitate communication and coordination between agents (i.e., neural models). Symbolic reasoning mediates interactions, enforces constraints, and ensures alignment with predefined rules, while neural components adapt and learn from collective behaviors. This interplay enables robust and scalable problem-solving in complex, multi-agent environments. Belle et al.  \cite{belle2023neuro} explored how the combination of symbolic reasoning and agents can enable the development of advanced systems that are closer to human-like intelligence. They discusses how symbolic reasoning can mediate communication between agents, ensuring that they adhere to predefined rules while allowing the neural components to learn and adapt from collective behaviors. This directly aligns with the fibring paradigm, where multiple NNs are interconnected via a symbolic fibring function, enabling them to collaborate and share information in a structured manner.

Similarly, the recent DeepSeek-R1  \cite{guo2025deepseek} framework employs a MoE architecture to enhance reasoning capabilities in large-scale AI systems. DeepSeek’s MoE approach activates only a subset of its parameters for each task, mimicking a team of specialized experts. These experts coordinate effectively using reinforcement learning rewards and symbolic constraints, enabling efficient resource utilization while ensuring adherence to reasoning rules. The symbolic constraints act as an intermediary layer, guiding the interactions between experts in a structured manner, aligning their individual outputs to form a cohesive solution.

Likewise, Mixtral 8x7B \cite{jiang2024mixtral} employs a sparse mixture-of-experts (SMoE) framework, where each layer selects specific expert groups to process input tokens. This architecture not only reduces computational costs but also ensures that the model specializes in handling different tasks through expert routing. Mixtral’s ability to adaptively select experts for tasks requiring mathematical reasoning or multilingual understanding exemplifies how MoE-based systems achieve scalability and specialization while maintaining efficiency. The symbolic mediator within Mixtral ensures that expert selection follows a structured process governed by logical rules, promoting an orderly exchange of information between the experts while adhering to predefined symbolic constraints.

\section{Evaluation of NSAI Architectures}

Ensuring the reliability and practical applicability of NASAI architectures requires a systematic evaluation across multiple well-defined criteria. Such an evaluation not only identifies the strengths and limitations of the architectures but also fosters trust among stakeholders by emphasizing interpretability, transparency, and robustness—qualities essential in domains such as healthcare, finance, and autonomous systems. Moreover, a rigorous assessment provides benchmarks that can stimulate the development of next-generation models. The following sections delineate the key criteria for evaluating NSAI architectures. 

\subsection{Core Criteria}
The evaluation framework for NSAI architectures is built upon several fundamental criteria: generalization, scalability, data efficiency, reasoning, robustness, transferability, and interpretability. Each criterion is elaborated below.\\

\noindent \textbf{Generalization:}
Generalization is defined as the capability of a model to extend its learned representations beyond the training dataset to perform effectively in novel or unforeseen situations. This criterion is evaluated based on:

\begin{itemize}
    \item[--] \textit{Out-of-distribution (OOD) performance}: The ability to maintain performance on data that deviate from the training distribution.
    \item[--] \textit{Contextual flexibility}: The capacity to adapt seamlessly to changes in context or domain with minimal retraining.
    \item[--] \textit{Relational accuracy}: The capacity to identify and exploit relevant relationships in  data while mitigating the influence of spurious correlations.
\end{itemize}

\noindent \textbf{Scalability:}
Scalability assesses the performance of NSAI architecture under
increasing data volumes or computational demands. A scalable system should remain efficient and effective as it scales. Key aspects include:

\begin{itemize}
    \item[--] \textit{Large-scale adaptation}: The ability to process and derive insights from massive datasets.
    \item[--] \textit{Hardware efficiency}: Optimal utilization of available computational resources, enabling operation on both low-resource devices and high-performance infrastructures.
    \item[--] \textit{Complexity management}: The ability to accommodate increased architectural complexity without compromising speed or deployment feasability.
\end{itemize}

\noindent \textbf{Data Efficiency:}
Data efficiency measures how effectively an NSAI model learns from limited data, an important consideration in scenarios where labeled data are scarce or expensive to obtain. This criterion encompasses: 

\begin{itemize}
    \item[--] \textit{Data reduction}: Achieving high performance with a reduced amount of training data.
    \item[--] \textit{Data optimization}: Maximizing the utility of available data (both labeled and unlabeled), potentially through semi-supervised learning techniques.
    \item[--] \textit{Incremental adaptability}: The capacity to incorporate new data progressively without undergoing complete retraining.
\end{itemize}

\noindent \textbf{Reasoning:}
Reasoning reflects the model's ability to analyze data, extract insights, and draw logical conclusions. This criterion underscores the unique advantage of NSAI architectures, which combine neural learning with symbolic reasoning. This criterion evaluates:

\begin{itemize}
    \item[--] \textit{Logical reasoning}: The systematic application of explicit rules to derive precise and consistent inferences. 
    \item[--] \textit{Relational understanding}: The comprehension of complex relationships between entities within the data.
    \item[--] \textit{Cognitive versatility}: The integration of various reasoning paradigms (e.g., deductive, inductive, and abductive reasoning) to tackle diverse challenges. 
\end{itemize}

\paragraph{Robustness:}
Robustness measures the system’s reliability and resilience to disruptions, including noisy data, adversarial inputs, or dynamic environments. The evaluation considers:

\begin{itemize}
    \item[--] \textit{Resilience to perturbations/anomalies}: The ability to sustain stable performance despite the presence of noise or adversarial data.
    \item[--] \textit{Adaptive resilience}: The maintenance of functionality under changing or unpredictable conditions.
    \item[--] \textit{Bias resilience}: The effectiveness in detecting and correcting biases to ensure fairness and accuracy in predictions.
\end{itemize}

\noindent \textbf{Transferability:}
Transferability assesses the model’s ability in applying learned knowledge to new contexts, domains, or tasks. This is essential for reducing the effort and time required for model adaptation. Its evaluation involves:

\begin{itemize}
    \item[--] \textit{Multi-domain adaptation}: The capacity to generalize across diverse domains with minimal modifications.
    \item[--] \textit{Multi-task learning}: The capability to handle multiple tasks simultaneously through shared knowledge representations.
    \item[--] \textit{Personalization}: The adaptability of the model to meet specific user or application requirements with limited additional effort.
\end{itemize}

\noindent \textbf{}{Interpretability:}
Interpretability evaluates the model’s ability to explain its decisions, ensuring transparency and trust in NSAI systems. This criterion assesses:

\begin{itemize}
    \item[--] \textit{Transparency}: The clarity with which the internal mechanisms and decision processes of the model are revealed.
    \item[--] \textit{Explanation}: The ability to provide comprehensible justifications for predictions or decisions.
    \item[--] \textit{Traceability}: The capability to reconstruct the sequence of operations and factors that contributed to a given outcome.
\end{itemize}

\noindent By systematicaly addressing these criteria, researchers and practitioners can ensure that NSAI architectures are not only scientifically rigorous but also practical, adaptable, and ready for real-world applications. This evaluation framework not only facilitates continuous improvement and innovation but also supports the broad adoption of NSAI systems across various industries and application domains.

\subsection{Evaluation Methodology}

\noindent The evaluation of NSAI architectures was conducted using a systematic approach to ensure a robust and transparent assessment of their performance across multiple criteria. This process relied on three key sources: scientific literature, empirical findings, and an analysis of the design principles underlying each architecture. \textbf{Table~\ref{tab:references}} summarizes the relevant research works associated with the identified NSAI architectures in Section 3. 
 The scientific literature served as the primary source of qualitative insights, offering detailed analyses of the strengths and limitations of various architectures. Foundational research and state-of-the-art studies provided evidence of performance in areas such as scalability, reasoning, and interpretability, helping to guide the evaluation. Additionally, empirical results from experimental studies and benchmarks offered quantitative data, enabling objective comparisons across architectures. Metrics such as accuracy, adaptability, and efficiency were particularly valuable in validating the claims made in research papers. The design principles of each technology were also considered to understand how neural and symbolic components were integrated. This analysis provided insights into the inherent capabilities and constraints of each architecture, such as its suitability for handling complex reasoning tasks, scalability to large datasets, or adaptability to dynamic environments.

\vspace*{0.5cm}

\noindent For each criterion, the ratings were assigned as follows:

\begin{itemize}
    \item \textit{High:} Awarded to architectures that consistently demonstrated exceptional performance across multiple studies and benchmarks, showcasing clear advantages in the specific criterion.
    \item \textit{Medium:} Assigned to architectures with satisfactory performance, excelling in certain aspects but with notable limitations in others.
    \item \textit{Low:} Given to architectures with significant weaknesses, such as inconsistent results or an inability to effectively address the criterion.
\end{itemize}

 By combining insights from literature, empirical findings, and design analysis, this methodology ensures a balanced and evidence-based evaluation. It provides a clear understanding of the strengths and weaknesses of each architecture, enabling meaningful comparisons and guiding future advancements in NSAI research and applications.

\begin{table}[h]
    \centering
        \caption{Set of relevant published NSAI architectures considered in the proposed study.}

    \begin{tabular}{|l|p{10cm}|}
        \hline
        \textbf{Architecture} & \textbf{References} \\
        \hline
        \textit{Symbolic $\to$ Neuro $\to$ Symbolic} & \cite{kouris2021abstractive}, \cite{sutherland2019leveraging}, \cite{gu2019local}, \cite{cui2021sememes}, \cite{xu2019relation}, \cite{cowen2019neural}, \cite{bounabi2021new}, \cite{es2021sentence}, \cite{lima2019impact}, \cite{zhou2021relation}, \cite{gong2020hierarchical}, \cite{tato2019hybrid}, \cite{langton2021applied}, \cite{bracsoveanu2019semantic}, \cite{pinhanez2021using}, \cite{dehua2021bdcn}, \cite{fazlic2019novel}, \cite{d2019team}, \cite{ayyanar2019causal}, \cite{hu2021dialoguecrn}, \cite{chen2020question}, \cite{manda2020automated}, \cite{honda2019question}, \cite{schon2019corg}, \cite{amin2019cases} \\
        \hline
        \textit{Neuro[Symbolic]} & \cite{heule2016solving}, \cite{madan2021fast} \\
        \hline
        \textit{Symbolic[Neuro]} & \cite{silver2016mastering}, \cite{chen2021web}, \cite{chen2021neurallog}, \cite{pacheco2021modeling}, \cite{chaturvedi2019fuzzy}, \cite{qin2021neural} \\
        \hline
        \textit{Neuro $|$ Symbolic} & \cite{mao2019neuro}, \cite{yao2018learning}, \cite{shi2021neural}, \cite{vskrlj2021autobot}, \cite{wang2021variational}, \cite{lemos2020neural}, \cite{huang2019attentive} \\
        \hline
        \textit{Neuro $\to$ Symbolic $\leftarrow$ Neuro} & \cite{das2021case}, \cite{garcez2004fibring}, \cite{belle2023neuro}, \cite{guo2025deepseek}, \cite{jiang2024mixtral}, \cite{guo2024large}, \cite{maldonado2024multi}, \cite{he2024mixturemillionexperts}, \cite{lo2024closerlookmixtureofexpertslarge} \\
        \hline
        \textit{Neuro:Symbolic $\to$ Neuro} & \cite{lample2019deep}, \cite{yabloko2020ethan}, \cite{zhou2020temporal}, \cite{saveleva2021graph}, \cite{gupta2021neuro}, \cite{demeter2020just}, \cite{jiang2021lnn}, \cite{kogkalidis2020neural}, \cite{zhang2021noahqa}, \cite{sen2020learning}, \cite{huo2019graph}, \cite{jiang2020medical}, \cite{liu2021heterogeneous}, \cite{chaudhury2021neuro}, \cite{verga2020facts}, \cite{socher2013reasoning}\\
        \hline
        \textit{Neuro\textsubscript{Symbolic\textsubscript{Loss}}} & \cite{serafini2016logic}, \cite{raissi2019physics}, \cite{chen2020mapping}, \cite{graziani2019jointly}, \cite{altszyler2020zero}, \cite{hussain2018semi} \\
        \hline
        \textit{Neuro\textsubscript{Symbolic\textsubscript{Neuro}}} & \cite{smolensky2016basicreasoningtensorproduct} \cite{smolensky1990tensor} \\
        \hline
    \end{tabular}
    \label{tab:references}
\end{table}

\subsection{Results and Discussion}
\begin{figure}
    \centering
    \includegraphics[width=1\linewidth]{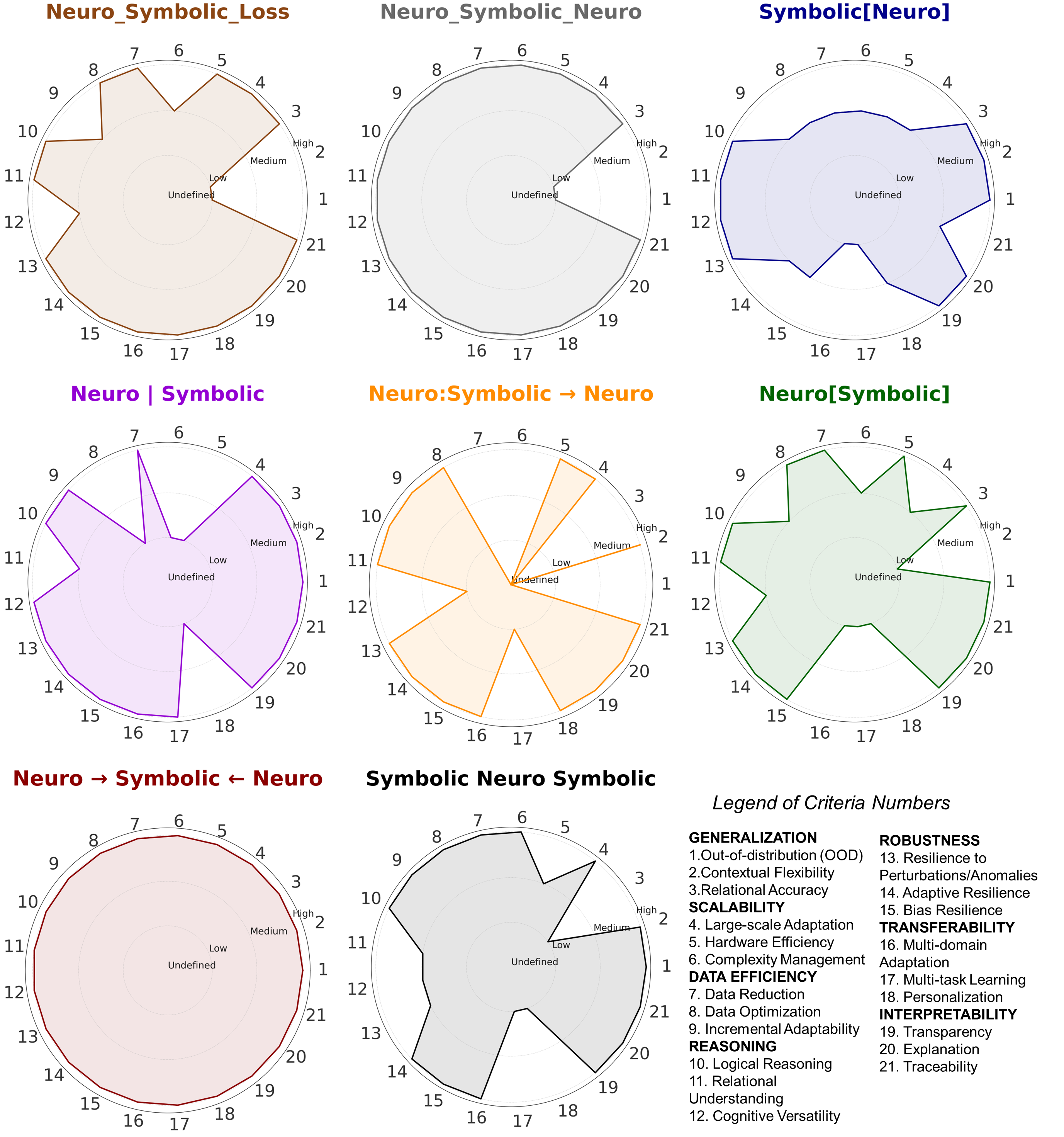}
    \caption{Comparison of NSAI architectures based on various criteria and sub-criteria.}
    \label{comparison}
\end{figure}

\noindent \textbf{Figure~\ref{comparison}}  provides a comparative analysis of various NSAI architectures across seven main evaluation criteria and their respective sub-criteria. This comprehensive evaluation highlights the strengths and weaknesses of each architecture, offering insights into their performance, adaptability, and interpretability.

For example, under the ``generalization" criterion, \textit{Neuro $\to$ Symbolic $\leftarrow$ Neuro} and \textit{Neuro $|$ Symbolic} perform well in generalization scenarios, demonstrating strong generalization capabilities, particularly in handling relational accuracy, making it suitable for complex, real-world applications. However, \textit{Neuro\textsubscript{Symbolic\textsubscript{Loss}}} and \textit{Neuro\textsubscript{Symbolic\textsubscript{Neuro}}} demonstrates notable shortcomings in continuous flexibility and OOD generalization, highlighting its difficulty in adapting to dynamic and evolving contexts without the need for extensive retraining. As for the ``scalability" criterion, \textit{Neuro $\to$ Symbolic $\leftarrow$ Neuro} and \textit{Neuro\textsubscript{Symbolic\textsubscript{Neuro}}} excel across all sub-criteria, including large-scale adaptation and hardware efficiency, demonstrating their capacity to handle industrial-scale applications. Conversely, \textit{Symbolic[Neuro]} achieves only medium performance in scalability, reflecting challenges in balancing its rule-based reasoning with the demands of large-scale or resource-intensive tasks. In particular, \textit{Neuro $|$ Symbolic}, rated low, struggles to maintain efficiency and adaptability when scaling to more complex systems, highlighting a need for improved coordination between its neural and symbolic components. 

\vspace*{0.5cm}

In terms of ``data efficiency", architectures such as \textit{Neuro $\to$ Symbolic $\leftarrow$ Neuro}, \textit{Symbolic Neuro Symbolic}, and \textit{Neuro\textsubscript{Symbolic\textsubscript{Neuro}}} consistently achieve high ratings, excelling in both data reduction and optimization. This indicates their ability to learn effectively with limited data. However, \textit{Symbolic[Neuro]} demonstrates only medium adaptability when incorporating incremental data updates. When evaluating the ``Reasoning" criterion, architectures such as \textit{Symbolic[Neuro]}, \textit{Neuro $\to$ Symbolic $\leftarrow$ Neuro}, and \textit{Neuro\textsubscript{Symbolic\textsubscript{Neuro}}} show strong capabilities in logical reasoning and relational understanding. However, \textit{Neuro:Symbolic $\to$ Neuro} displays lower versatility in combining diverse reasoning methods, reflecting limitations in solving complex problems. For ``Robustness", most architectures perform well, demonstrating high resilience to perturbations and effective bias handling. However, \textit{Symbolic[Neuro]} and \textit{Symbolic Neuro Symbolic} architectures exhibit weaknesses in adapting to dynamic environments and mitigating biases effectively. 

\vspace*{0.5cm}

Regarding ``Transferability", architectures like \textit{Neuro $\to$ Symbolic $\leftarrow$ Neuro}, \textit{Neuro\textsubscript{Symbolic\textsubscript{Loss}}}, and \textit{Neuro\textsubscript{Symbolic\textsubscript{Neuro}}} excel in multi-task learning and multi-domain adaptation, enabling effective reuse of knowledge across domains. In contrast, \textit{Symbolic Neuro Symbolic}, \textit{Neuro:Symbolic $\to$ Neuro}, and nested architectures demonstrate lower adaptability to personalized applications.
Lastly, in ``Interpretability", most architectures perform well, achieving high marks in transparency and traceability. \textit{Symbolic[Neuro]} also achieves commendable results in this criterion, demonstrating its ability to explain decisions effectively, which is essential for sensitive applications like healthcare and finance.

\vspace*{0.5cm}

Overall, the \textit{Neuro $\to$ Symbolic $\leftarrow$ Neuro} architecture emerges as the best-performing model, consistently achieving high ratings across all criteria. Its exceptional performance in generalization, scalability, and interpretability makes it highly suitable for real-world applications that demand reliability, adaptability, and transparency. While other architectures also perform well in specific areas, the versatility and robustness of \textit{Neuro $\to$ Symbolic $\leftarrow$ Neuro} set it apart as the most balanced and capable solution. This conclusion aligns with findings in the state of the art, which highlight the effectiveness of \textit{Neuro $\to$ Symbolic $\leftarrow$ Neuro} architectures in leveraging advanced AI technologies, such as multi-agent systems. Multi-agent systems are well-documented for their robustness, particularly in dynamic and distributed environments, where their ability to coordinate, adapt, and reason collectively enables superior performance. For instance, Subramanian et al. \cite{subramanian2024neuro} demonstrated that incorporating neuro-symbolic approaches into multi-agent RL enhances both interpretability and probabilistic decision-making. This makes such systems highly robust in environments with partial observability or uncertainties. Similarly, Keren et al.  \cite{keren2021collaboration} highlighted that collaboration among agents in multi-agent frameworks promotes group resilience, enabling these systems to adapt effectively to dynamic or adversarial conditions. These attributes are particularly valuable in \textit{Neuro $\to$ Symbolic $\leftarrow$ Neuro} architectures, as they address the critical need for transparency and robustness in complex real-world applications.



\section{Conclusion}
This study evaluates several NSAI architectures against a comprehensive set of criteria, including generalization, scalability, data efficiency, reasoning, robustness, transferability, and interpretability. The results highlight the strengths and weaknesses of each architecture, offering valuable insights into their capabilities for real-world applications. Among the architectures investigated, \textit{Neuro $\to$ Symbolic $\leftarrow$ Neuro} emerges as the most balanced and robust solution. It consistently demonstrates superior performance across multiple criteria, excelling in generalization, scalability, and interpretability. These results align with recent advancements in the field, which emphasize the role of multi-agent systems in enhancing robustness and adaptability. As shown in recent studies, multi-agent frameworks, when integrated with neuro-symbolic methods, provide significant advantages in handling uncertainty, fostering collaboration, and maintaining resilience in dynamic environments. This integration not only enables better decision-making but also ensures transparency and traceability, which are critical for sensitive applications.  Moreover, its ability to leverage advanced AI technologies, such as multi-agent systems, positions \textit{Neuro $\to$ Symbolic $\leftarrow$ Neuro} as a leading candidate for addressing the demands of generative AI applications.

\vspace*{0.5cm}

Future work will be focused on exploring the scalability of this architecture in even larger and more diverse environments. Additionally, advancing the integration of symbolic reasoning within multi-agent systems may further enhance their robustness and cognitive versatility. As the field evolves, \textit{Neuro $\to$ Symbolic $\leftarrow$ Neuro} architectures are likely to remain at the forefront of innovation, offering practical and scientifically grounded solutions to the most pressing challenges in AI.

\section*{CRediT authorship contribution statement}
\textbf{Oualid Bougzime:} Writing – original draft, Methodology, Investigation. \textbf{Samir Jabbar:} Writing – original draft, Methodology, Investigation. \textbf{Christophe Cruz:} Writing – review \& editing, Methodology, Supervision. \textbf{Fr\'ed\'eric Demoly:} Writing –
review \& editing, Methodology, Supervision, Funding acquisition, Project administration.

\section*{Declaration of competing interest}
The authors declare that they have no known competing financial interests or personal relationships that could have appeared to influence the work reported in this paper.

\section*{Acknowledgements}
This research was funded by the IUF, Innovation Chair on 4D Printing, the French National Research Agency under the “France 2030 Initiative” and the “DIADEM Program”, grant number 22-PEXD-0016 (“ARTEMIS”).

\bibliographystyle{unsrt}

\end{document}